\documentclass{article}
\usepackage{acra}
\usepackage{graphicx}
\usepackage{multirow}
\usepackage{adjustbox}
\usepackage{amsmath}
\usepackage{url}
\usepackage{hyperref}
\usepackage{array}
\usepackage{subcaption}

\title{Look how they have grown: Non-destructive Leaf Detection and Size Estimation of Tomato Plants for 3D Growth Monitoring}
\author{Yuning Xing$^{*}$, Dexter Pham, Henry Williams, David Smith, Ho Seok Ahn\\ \textbf{JongYoon Lim, Bruce A. MacDonald, Mahla Nejati$^{**}$}\\
Centre for Automation and Robotic Engineering Science\\ The University of Auckland, NZ\\
\{$^{*}$yxin683@aucklanduni.ac.nz, $^{**}$m.nejati@auckland.ac.nz\} }

\begin{document}
% Look_how_they_have_grown_Non-destructive_Leaf_Detection_and_Size_Estimation_of_Tomato_Plants_for_3D_Growth_Monitoring
\maketitle

\begin{abstract}
    Smart farming is a growing field as technology advances. Plant characteristics are crucial indicators for monitoring plant growth. Research has been done to estimate characteristics like leaf area index, leaf disease, and plant height. However, few methods have been applied to non-destructive measurements of leaf size. In this paper, an automated non-destructive imaged-based measuring system is presented, which uses 2D and 3D data obtained using a Zivid 3D camera, creating 3D virtual representations (digital twins) of the tomato plants. Leaves are detected from corresponding 2D RGB images and mapped to their 3D point cloud using the detected leaf masks, which then pass the leaf point cloud to the plane fitting algorithm to extract the leaf size to provide data for growth monitoring. The performance of the measurement platform has been measured through a comprehensive trial on real-world tomato plants with quantified performance metrics compared to ground truth measurements. Three tomato leaf and height datasets (including 50+ 3D point cloud files of tomato plants) were collected and open-sourced in this project. The proposed leaf size estimation method demonstrates an RMSE value of 4.47mm and an $R^2$ value of 0.87. The overall measurement system (leaf detection and size estimation algorithms combine) delivers an RMSE value of 8.13mm and an $R^2$ value of 0.899. 
\end{abstract}

\section{Introduction}
    % This introduction should be adjusted to either state this is work towards the idea of automating de-leafing or strengthening the justification for the growth monitoring side of things. Right now I am torn as to which works better as I am still unsure exactly what the end goal of this work is trying to acheive...what is going to happen if this system works? where will the data be applied?
    
    % Think we are definetly focusing on growth monitoring side of thing. The system was aimed to reduce human supervision as in greenhouses human inspection is required every week for every row.  Through automating this process, this system effectively reduce human error and inefficiency.  
    The New Zealand tomato industry generates more than \$130M a year with \$10M in exports from covered growing operations \cite{tomatoesNZ2022}.
    Following a similar global trend, the lack of reliable access to skilled labour is causing critical issues to productivity and quality as the industry expands.
    One of the primary challenges is maintaining skilled human operators for growth monitoring and de-leafing tomato plants. 
\begin{figure}[!htbp]
        \centering
        \includegraphics[width=\linewidth]{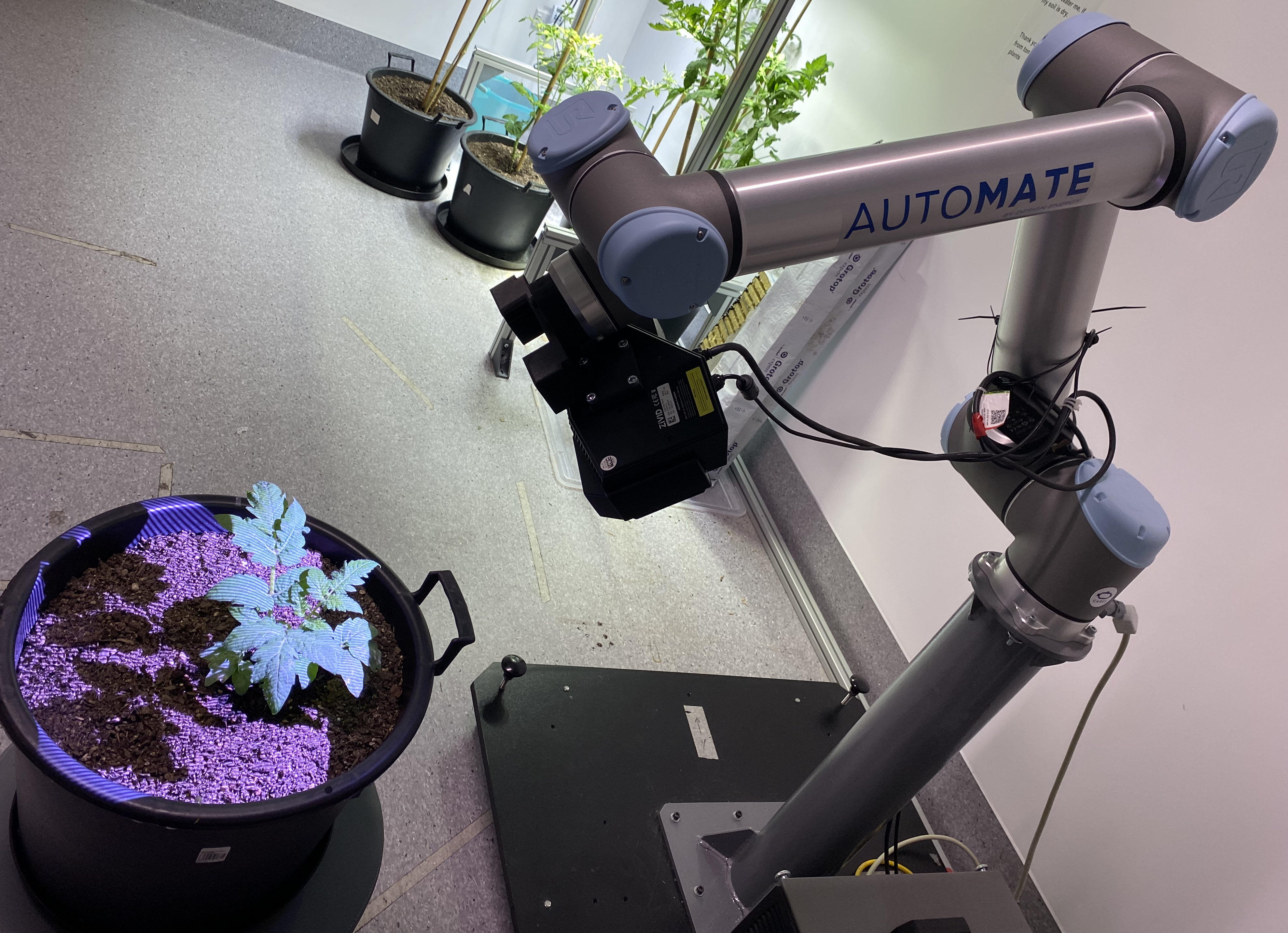}
        \caption{Robotic platform used to scan the tomato plants with attached Zivid camera.}
        \label{fig:scanning_rig}
    \end{figure}
    Growth monitoring is key to the general management of growing plants and is helpful for early disease detection. 
    The growth rate can be tracked through the leaf length and width of the plant.
    This information feeds into disease management protocols for the given orchard. 
    
    De-leafing the tomato plant is a vital part of the growing cycle to foster the growth of quality tomatoes. 
    As the plant grows, pruning out any crossing, crowded, damaged, or diseased stems and foliage helps to keep the plant open, airy, and free of pests and diseases.
    Removing tomato plant leaves that grow beneath the flower sets will also send more energy to fruit formation, producing higher-quality tomatoes faster.
    Current growth monitoring and de-leafing approaches are conducted manually, ideally once a week. This task requires skilled and detail attentive workers' full attention and is highly repetitive for the endless rows in a greenhouse.
    
    Our work aims to develop a robotic platform that automatically monitors tomato plants as they grow. Due to the nature of growth monitoring, the approach has to be non-destructive. In light of this, as the growth of each measurement is calculated and compared against each other, the absolute truth of size is less vital, reducing the human error involved in obtaining the data. The curvature of the leaf is not measured in this work because it does not affect the applied methods, as the current system produces a size estimation of the leaf, which can be used to monitor growth or decrease in size. Thus, the outputs of this system can be used to identify such abnormal changes when appropriate.
    
    This paper presents the initial design, implementation, and evaluation details of the vision system of this robotic platform.
    The platform consists of a UR10 robotic arm and a Zivid Camera, enabling it to map the plant's structure and measure its key features: leaf size and height.
    Figure \ref{fig:scanning_rig} presents a photo of the scanning platform prototype.
    % Section \ref{sec:videos} provides videos of it operating.
    
    The performance of the monitoring platform has been measured through a comprehensive trial on real-world tomato plants with quantified performance metrics compared to ground truth measurements. Mask R-CNN was used to detect the leaves. Two types of non-destructive manual measurements were used: strength measurement (the entire length of the leaf) and natural measurement (measuring the leaf in its normal state). Three ways of leaf size estimation were tested and evaluated.
    The results of these field trials are presented, and the implications for further platform development are discussed.
     
\section{Related Works}
    %Example of how to cite papers on our past work with kiwifruit \cite{williams2019robotic,williams2020autonomous,williams2020improvements,williams2021evaluating}.
    %\Mahla's comment on how to \cite{}, Find the paper on Mendelay. Then, find the citation key on the details part of each paper(right side of the screen). Add the citation like \cite{Citation Key}. If the citation key does not show on details go to tools->options->documents Details, select the type of document type and then tick the citation key. The citation key usually is the name and year of the publication. If the citation key is the same as the another publication, add a at the end of the year to the citation key like Zhang2012a
    % or search the paper on google scholar cite->bibTex
    This section will review a series of literature to better understand the current state of methods and algorithms presented on leaf detection and measurement. First, the existing leaf detection approach will be explored. Then, leaf measurement techniques and approaches will be surveyed.
    \subsection{Detection}
    Currently, there are several approaches to detecting plant components in images for various applications \cite{williams2019robotic,Nejati2019,williams2021evaluating}. Two main approaches for extracting a leaf from an image are based on image segmentation or machine learning methods. 

    In the first approach - image segmentation - image processing techniques such as colour thresholding, GrabCut, Watershed, and Random Walker are commonly implemented. \cite{chen2017plant} utilised different colour spaces, such as CIELAB, and HSI/HSV, to deal with different lighting conditions. \cite{yeh2014automated,anantrasirichai2017automatic} employed the GrabCut algorithm to detect and measure a leaf's area, width and length. Since GrabCut requires a precision marker for better segmentation results, \cite{anantrasirichai2017automatic} developed markers using the information given by the intensity and texture of the image. Watershed was employed to segment occluded cotton leaves. 

    For the machine learning-based approach, Deep Neural Networks have been commonly proposed in various literature.
    Yang et al. \cite{yang2020leaf} compared a Mask R-CNN machine learning model with GrabCut and Otsu segmentation algorithms for a leaf segmentation task with a complicated background. The misclassification error of each technique was 1.15\%, 28.74\%, and 29.85\%, respectively. However, the machine learning approach was trained using more than 4000 images of 15 plant species on an NVIDIA Tesla P4 GPU. In contrast, the image segmentation approach requires neither a large dataset nor special hardware.

    Giuffrida et al. \cite{giuffrida2016learning} proposed a machine learning-based approach to estimate the number of leaves from top-view images. This approach achieved the absolute-difference-in-object-count $|DiC|$ values of 1.27 and 1.36 on the Leaf Counting Challenge 2015 (LCC 2015) dataset A1 \cite{o2010linking}, and A3 \cite{scharr2014annotated}, which are significantly smaller than the $|DiC|$ values of 2.2 and 2.8 from an image segmentation approach (3D histogram) presented by Pape et al. \cite{pape20143} for the Leaf Segmentation Challenge 2014. However, for dataset A2 \cite{o2010linking}, the machine learning approach only achieved the $|DiC|$ value of 2.44 compared to 1.2 from the 3D histogram approach. The reason was that both datasets A1 and A3 have more training data than A2. 
    
    Compared to machine learning  techniques, leaf segmentation using image processing techniques offer a simple implementation, as it requires neither a large dataset for the training process nor special computer hardware, which are common burdens of machine learning techniques. However, machine learning techniques provide higher performance in detection; According to \cite{yang2020leaf}, the misclassification error of the DNN was the lowest compared to the GrabCut and Otsu thresholding techniques.

    %Although machine learning approaches tend to produce more desirable results, they require a large dataset for the training process. 

    Furthermore, to our best knowledge, there exists no publicly available dataset for tomato plants at the time of this research. This is a challenge for this research to explore.

    \subsection{Measurement}
        Firstly, to help decide what sensor data should be used for measuring, the suitability of 2D data and 3D data will be compared. Then, feature-based and machine learning algorithms used in this field will be discussed.
         
        \subsubsection{2D data versus 3D data}
        2D RGB data can be used for estimating plant height and leaves through various methods. Gupta et al. \cite{gupta2022image} achieved a root mean square error of 0.003 – 0.006m using RGB images obtained from limited camera resolution using a Logitech Webcam Pro 9000, the algorithm used will be explained further in the image processing algorithms section. Schrader et al. \cite{schrader2017leaf} created a smartphone application that can estimate the leaf area using the camera from a phone to obtain RGB images; the results are promising with a precision of 0.981 - 0.992 given the required indoor environment, and lighting condition met. Although 2D data can be used to produce desirable results, it has its limitations of ideally perfectly setting up plants and lighting for the algorithms to function correctly.
        
        3D RGBD data or data point cloud is widely used to extract depth information. Wang et al. \cite{wang2014size} showed that depth image obtained indoors using a Kinect RGBD camera with a machine vision system is more robust and accurate with an RMSE of 3.4mm and $R^2 = 0.89$, comparing the method using colour images. Another approach that only RGBD data can use is depth segmentation from a top view; when obtaining an RGBD image from the top of the plant, each layer of the leaves can be seen due to its height difference \cite{chene2012use}. The data can then be used for size estimation using image processing algorithms. Zhang et al. \cite{zhang20183d} have tested and found that more accurate results(RMSE: $0.12 m^2$, $R^2= 0.98$) were produced when estimating growth parameters from 3D models of plants created using multi-view images (collected using digital single lens reflex  SLR  camera  Canon - EOS  Kiss  X7).
        
        As discussed above, 2D and 3D data have their strengths and limitations when working with other plants. Thus, in our research, 2D and 3D data will be examined with different methods when detecting and measuring tomato leaf and plant height, as tomato plants have slightly different leaves and plant structures. Images will be obtained from different angles to be compared and evaluated.

    \subsubsection{Feature-based leaf size measurement}
        In 1997, Nyakwende et al. \cite{nyakwende1997non} achieved a high correlation value for measuring leaf area in tomato plants; the value was as high as $R^2 = 0.992$ using a $572 * 386$ pixels CCD video camera and mathematical image processing algorithm. This non-destructive method requires a multiple-angle view of the plant, controlled lighting conditions,Do hen, this indicates that with our current camera and image processing technologies, a more robust and accurate result can be achieved when measuring the individual leaf size. Similarly, Gong et al. \cite{gong2013handheld} developed a semi-automatic handheld device (HHD) which achieved a 0.455\% maximum error compared to a leaf area meter. The algorithm requires a reference square to extract the pixel-to-actual area ratio, the leaf area is then calculated using mathematical equations. A similar pixel to the actual area ratio method will be tested in our research. However, this method requests the leaf to be nicely presented when obtaining the data, which is challenging in our scenario. Although the algorithm is unsuitable for directly producing the result, it could be combined with machine learning algorithms to improve the accuracy. 
        
       An Android application was developed to measure leaf area using a phone camera and a margin detection algorithm \cite{schrader2017leaf}, which achieved a high accuracy(Precision = 0.981 - 0.992) comparable with commercial software. The margin detection algorithm is robust and designed for challenging outdoor conditions. Area measurement has three steps: margin detection, pixel count and reference object comparison. The acquired images are first processed by converting to grayscales, then increasing contrast to highlight margins; weak and strong margins are blurred and enhanced, respectively. Lastly, light gradients are calculated and displayed. Despite the claim of the algorithm being robust in field conditions, the requirement of a reference point adds a challenge for bulk scanning.
        
        3D modelling methods have also been proposed in various literature (\cite{zhang20183d}, \cite{miller20153d}, \cite{andujar2016using}). The procedure for 3D modelling includes: camera and lens calibration, obtaining point cloud, dense point cloud, mesh 3D polygonal model, 3D model texture and finally, DSM model is generated. \cite{zhang20183d} shows that although 3D modelling increase accuracy in some cases, some errors or noise occur during reconstruction, especially for leaf area due to overlapping leaves. Andújar et al. \cite{andujar2016using} generated a 3D model using a 3D point cloud from a depth video stream obtained using a Kinect RGBD camera. Although 3D modelling proves higher accuracy in most cases, the trade-off is worth considering for the extra steps needed for generating the 3D model. In addition, leaf size measurement is challenging due to overlapping and hidden leaf, which needs to be considered.

    \subsubsection{Machine Learning Algorithms}
        This section will look at some machine learning algorithms used in this field. 
        Few machine-learning techniques have been used to estimate leaf size and plant height. However, the relevant literature on estimation will be reviewed and evaluated for suitability for our project.
        
        Simple linear regression (SLR) has been used to estimate plant height \cite{liu2021estimation} to increase efficiency, effectiveness and accuracy compared to traditional manual measuring methods, which are destructive and time-consuming. A top-down view RGB-D image (collected using a Kinect RGBD camera) approach was taken in this research, with height calculated by depth value to the plant's highest point, subtracting the depth value of the mouth of the pot. Liu et al. claim that this method is reliable for its high prediction accuracy with a $R^2 = 0.72$; however, this correlation is weaker than the image processing algorithms presented in the previous section. Another method worth noting in this research is the use of the U-Net network for leaf segmentation. This method is potentially suitable for leaf size estimation, which will be considered in our research. 
        
         In \cite{mao2019comparison}, five machine learning regression algorithms were tested and compared for leaf area index estimation using Sentinel-2 imagery. These algorithms were evaluated by their accuracy, computational efficiency, and sensitivity to training sample size. The gradient boosting regression tree (GBRT) algorithm has the best accuracy of $R^2 = 0.864$. It is proven to be more robust with different training sample sizes, while the support vector regression (SVR) algorithm is the most efficient. GBRT may be tested and evaluated in our research to check if it is suitable for leaf size estimation.
         
        Partial least square regression (PLSR), v support vector regression (v-SVR) and gaussian process regression (GPR) was used to detect leaf disease. Different training sample sizes were tested again in this research \cite{ashourloo2016investigation} for result utilization. GPR is the most robust as it almost achieves the same degree of correlation despite the different training sample sizes. The training sample sizes will be considered when applying these algorithms in our research. 
        
        Therefore, machine learning algorithms have not yet been widely applied for measuring leaf size and plant height.

\section{Methods}
\begin{figure}[!htbp]
            \centering
             % {\includegraphics[width=5in, height=7cm]{images/systemd.jpg}}
            \includegraphics[width=\linewidth]{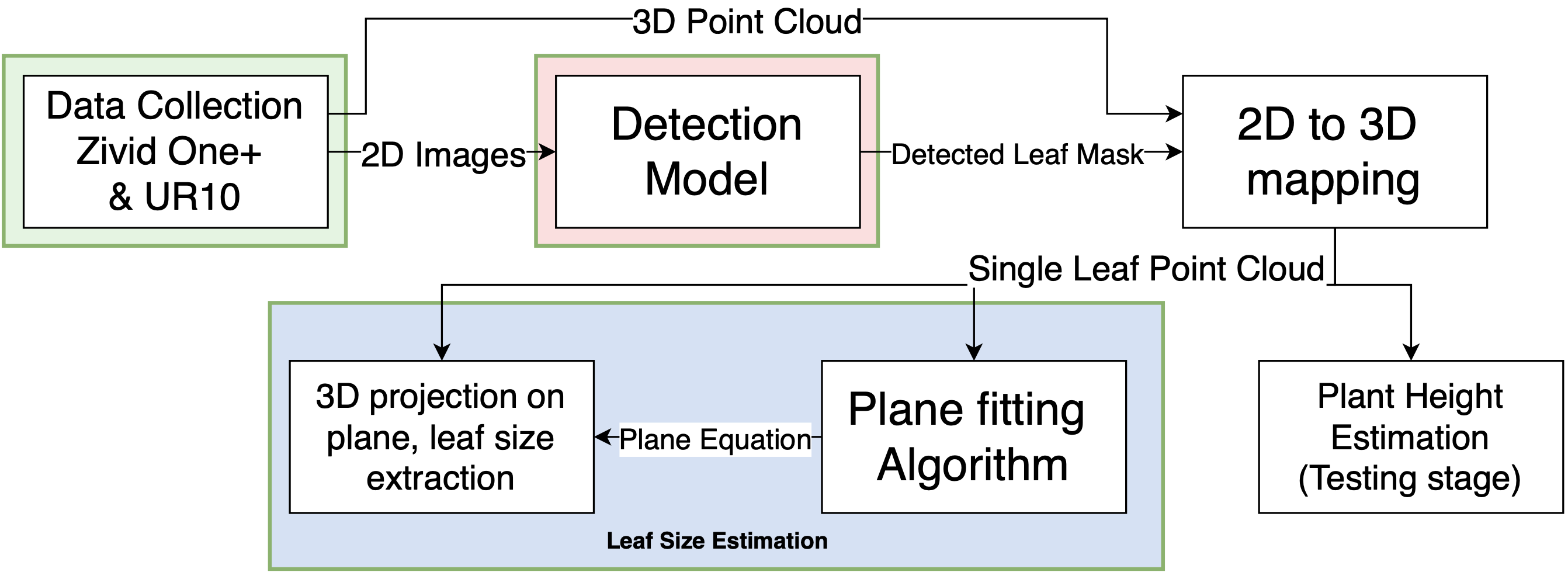}
            \caption{Proposed pipeline.}
            \label{fig:pipeline}
        \end{figure}

    The aim is to develop an image analysis system to detect the leaf and measure its size. The proposed approach goes through a three-step process of data collection, leaf detection and leaf size estimation, as shown in Figure \ref{fig:pipeline}. 
    
    % Brief about this section from Mahla: Add a section for data collection, you can mention the name of sensors that used and why? number of time that data collected, A link to the dataset, How much time did you spend for data collection, size of images and .... 
    
    \subsection{Data Collection}
        Due to the lack of public tomato leaf size and plant height dataset, the first step of this work was to collect high-quality data. A Zivid One Plus S 3D camera\footnote{\url{https://www.zivid.com/zivid-one-plus}} (Zivid, Oslo, Norway) and a pair of Basler acA2440-35uc USB 3.0 as stereo cameras (Basler AG) were mounted on a UR10 robot arm (Universal Robot, Doetinchem, The Netherlands) set up and was used to collect data in an indoor lab setting. These cameras are used for their high dynamic range and are suitable for indoor and outdoor conditions \cite{Nejati2019}. A Zivid camera was incorporated to obtain higher quality digital twins of the plants, as the stereo camera pair produces noisy point clouds. However, the combination of Zivid and stereo cameras provides a good dataset for the stereo-matching algorithm since Zivid data can be used as a ground truth. The Zivid camera has not previously been used for this purpose, but its accuracy is still above other depth cameras. By the time of this paper, 38 RGBD data(point cloud in ".ply" format and 2D images in ".png") were collected over seven different tomato plants; details of the dataset are shown in Table \ref{tab:datasetInfo}. 

\begin{table}[]
\caption{Dataset of leaf size and plant height measurement.}
\centering
\begin{adjustbox}{width=\linewidth}
\begin{tabular}{|ccccc|}
\hline
\multicolumn{1}{|c|}{Subsets} &
  \multicolumn{1}{c|}{Leaf-24} &
  \multicolumn{1}{c|}{Height-18} &
  \multicolumn{1}{c|}{\begin{tabular}[c]{@{}l@{}}Top-down\\ height\end{tabular}} &
  Misc** \\ \hline
\multicolumn{1}{|c|}{Leaf Size} &
  \multicolumn{1}{c|}{{\includegraphics[width=4mm,height=3mm]{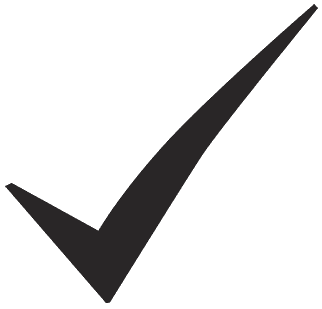}}} &
  \multicolumn{1}{c|}{} &
  \multicolumn{1}{c|}{} &
  {\includegraphics[width=4mm,height=3mm]{images/tick.png}} \\ \hline
\multicolumn{1}{|c|}{Plant Height} &
  \multicolumn{1}{c|}{} &
  \multicolumn{1}{c|}{{\includegraphics[width=4mm,height=3mm]{images/tick.png}}} &
  \multicolumn{1}{c|}{{\includegraphics[width=4mm,height=3mm]{images/tick.png}}} &
  {\includegraphics[width=4mm,height=3mm]{images/tick.png}} \\ \hline
\multicolumn{1}{|c|}{Number of Data*} &
  \multicolumn{1}{c|}{\begin{tabular}[c]{@{}l@{}}24\end{tabular}} &
  \multicolumn{1}{c|}{20} &
  \multicolumn{1}{c|}{8} &
  N/A \\ \hline
\multicolumn{1}{|c|}{Number of Pots} &
  \multicolumn{1}{c|}{4} &
  \multicolumn{1}{c|}{5} &
  \multicolumn{1}{c|}{4} &
  8+ \\ \hline
\multicolumn{1}{|c|}{Number of Scan} &
  \multicolumn{1}{c|}{4} &
  \multicolumn{1}{c|}{20} &
  \multicolumn{1}{c|}{8} &
  20+ \\ \hline
\multicolumn{1}{|c|}{Time per scan (Minutes)} &
  \multicolumn{1}{c|}{7} &
  \multicolumn{1}{c|}{7} &
  \multicolumn{1}{c|}{0.17} &
  5-60 \\ \hline
\multicolumn{5}{|c|}{\begin{tabular}[c]{@{}l@{}}*leaf size: number of leaves; plant height: number of pots; Leaf-24: 6 leaves/plot\\ ** Misc consist a range of 8 two-sided scans, each with 10 leaves measurements, \\ and many scans with no measurements which is good for detection training.
\end{tabular}} \\ \hline
\end{tabular}
\end{adjustbox}
\label{tab:datasetInfo}
\end{table}

The 3D point cloud files and available dataset measurements are open-sourced to the public. Please contact the corresponding author.
            
\begin{figure}[!htbp]
            \centering
            \includegraphics[width=70mm]{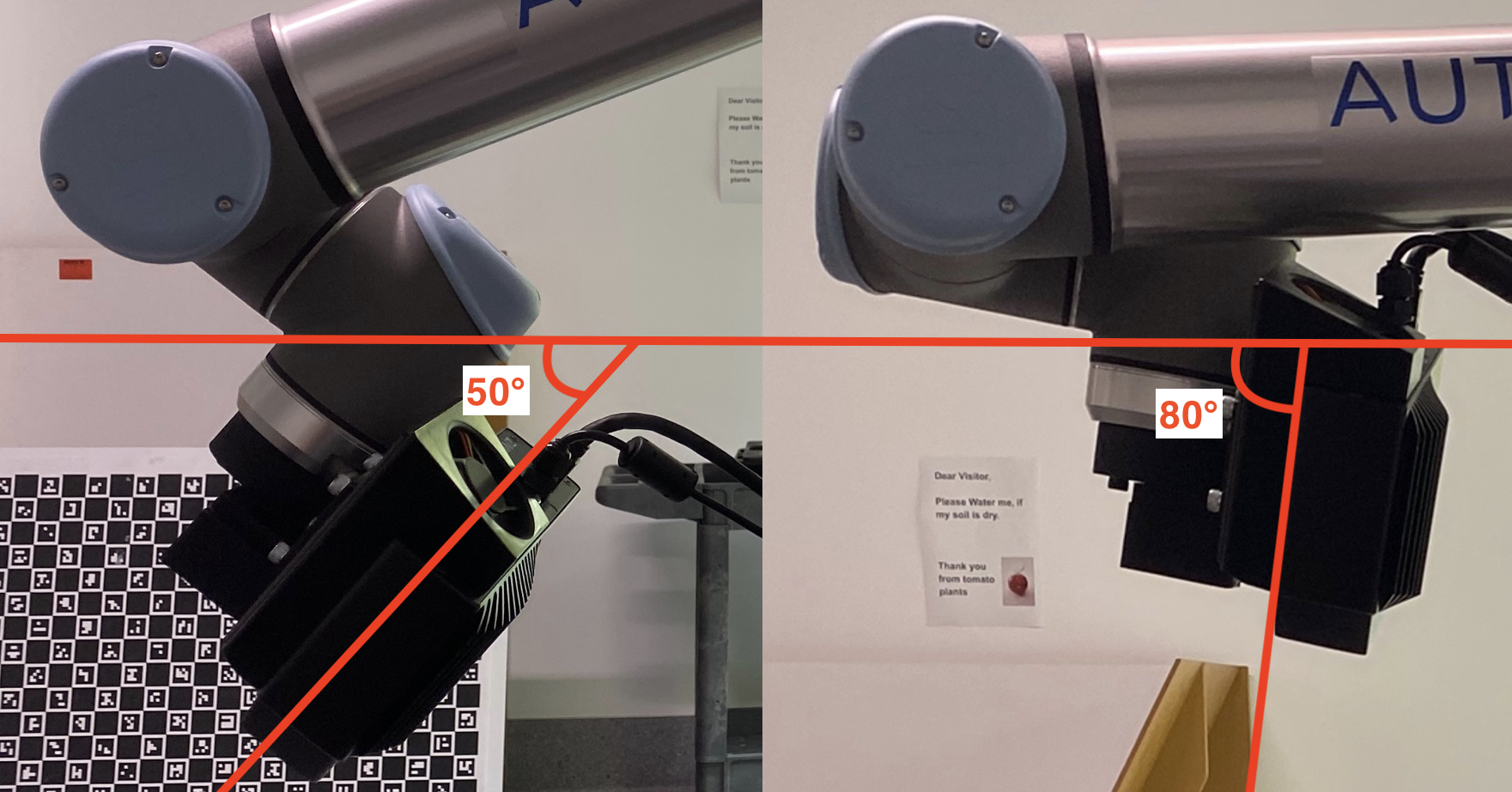}
            \caption{Angles used in dataset collection.}
            \label{fig:setupAngle}
        \end{figure}

        This dataset is primarily obtained using a 50-degree view, as shown in Figure \ref{fig:setupAngle}. This is the optimal angle found when testing the top-down view approach mentioned in the literature \cite{jiang2016high,chene2012use,zhang20183d}. This angle lets the camera obtain most leaf area, plant height and soil, thus achieving maximum information extraction for the measurements of leaf size and plant height. Plant height data have not been used in this research. Only using a single side scan instead of combining two or four sides of the scan decreases stitching error and increases data collection efficiency since the quality and data obtained from a single side scan are sufficient in providing data. The top-down 90 degrees view was tested and proven not ideal for plant height measurement as it can not scan the soil within the pot when the plant grows, and the leaf covers the pot and soil. This angle not only makes it impossible to extract the plant height but also decreases the amount of size leaf details the camera sees, leading to higher errors when automatically detecting these features. This complete top-down view also increases reflections from the background when scanning. Ninety degrees scan was also hard to achieve due to the camera and robot arm setup. Thus, 80 degrees view scans were obtained and tested as shown in Figure \ref{fig:setupAngle}. 
        %This also increases the approach's feasibility, as obtaining multiple view scans in the greenhouse environment is time-consuming and inefficient. 
        
    \subsection{Leaf Detection}
       Machine learning approaches often provide higher performance in detection. However, they also require a large annotated dataset to perform well. Our leaf detection leverages deep neural networks for reliable detection of the leaves using instance segmentation. The idea was to obtain 2D instance masks predicted by the deep learning network, which are useful and can be mapped later on into 3D for the leaf size estimation stage, as shown in Figure \ref{fig:pipeline}. This step utilises Detectron 2\footnote{\href{https://github.com/facebookresearch/detectron2}{https://github.com/facebookresearch/detectron2}} as an Instance Segmentation approach to detect the leaves in the 2D images \cite{wu2019detectron2}. 
        
        The leaf detection stage used 2D images, and the image size collected from the 2D cameras is 2464x2056 pixels. They were resized to 1333x1000 pixels to reduce the training time. There were a total of 75 annotated images consisting of 1491 annotated tomato leaves from different growth stages. As the dataset size is small and image labelling is time-consuming, often requiring experts in the field to conduct it correctly, some data augmentation techniques are required to increase the dataset size. A built-in transform feature in Detectron 2 was utilised for data augmentation. First, the input images were rotated in four orientations (90 degrees each). Then, they were flipped both horizontally and vertically. Finally, the images were transformed under various brightness and contrast before feeding to the Mask R-CNN network. The implemented model is a baseline R-50-FPN-3x Mask R-CNN network, which was trained for 6000 iterations using default settings in Detectron 2.
        
        % \begin{figure}[!htbp]
        %     \centering
        %     \includegraphics[width=\linewidth]{images/augmentationTraining.png}
        %     \caption{Data augmentation and training pipeline}
        %     \label{fig:augmentation}
        % \end{figure}

        % Example of how we discuss it relative to the fruitlet work - replace Fruitlet with leaf/leaves etc and fill in the right details
        % Images were captured from several orchards throughout a variety of lighting conditions.
        % N images were manually labelled to detect \textit{fruitlets} themselves and their \textit{calyx}, and \textit{stem}.
        % The calyx and stem are not used yet, future work will seek to determine the orientation of the fruitlet. 
        % Figure \ref{fig:fruitlet-detection} shows an example of these labelled features.
        % %TODO measure the performance of the detection model after training
        % The final training performance was...
        
        %TODO fill image
        % \begin{figure}
        %     \centering
        %     \includegraphics[width=\textwidth]{images/detection-example.png}
        %     \caption{Example of images labelled for Instance Segmentation showing detection of the fruitlets (colour), stem (colour), and calyx (colour).}
        %     \label{fig:fruitlet-detection}
        % \end{figure}
    
    \subsection{Leaf Size Measurement}
        Multiple depth image processing techniques have been tested and implemented to achieve automatic leaf size detection: points to actual area ratio, reference cube comparison and plane fitting. The first two techniques were tested in previous studies, and the third option is proposed in this paper. First, each technique was tested manually in 3D point cloud processing software (CloudCompare\footnote{\url{https://www.danielgm.net/cc/}}) and the promising testing results were implemented. 
        
        The first approach was evaluated by cropping out individual leaves in CloudCompare, which indicates the number of points associated with the leaf. Then, The number of points of each leaf is compared with the leaf size; however, no correlation was found between them, which proves this method unsuitable. Similarly, the reference cubes comparison method is unfit as it also relies on the number of points each leaf contains.

\begin{figure}[!htbp]
            \centering
            \includegraphics[width=\linewidth]{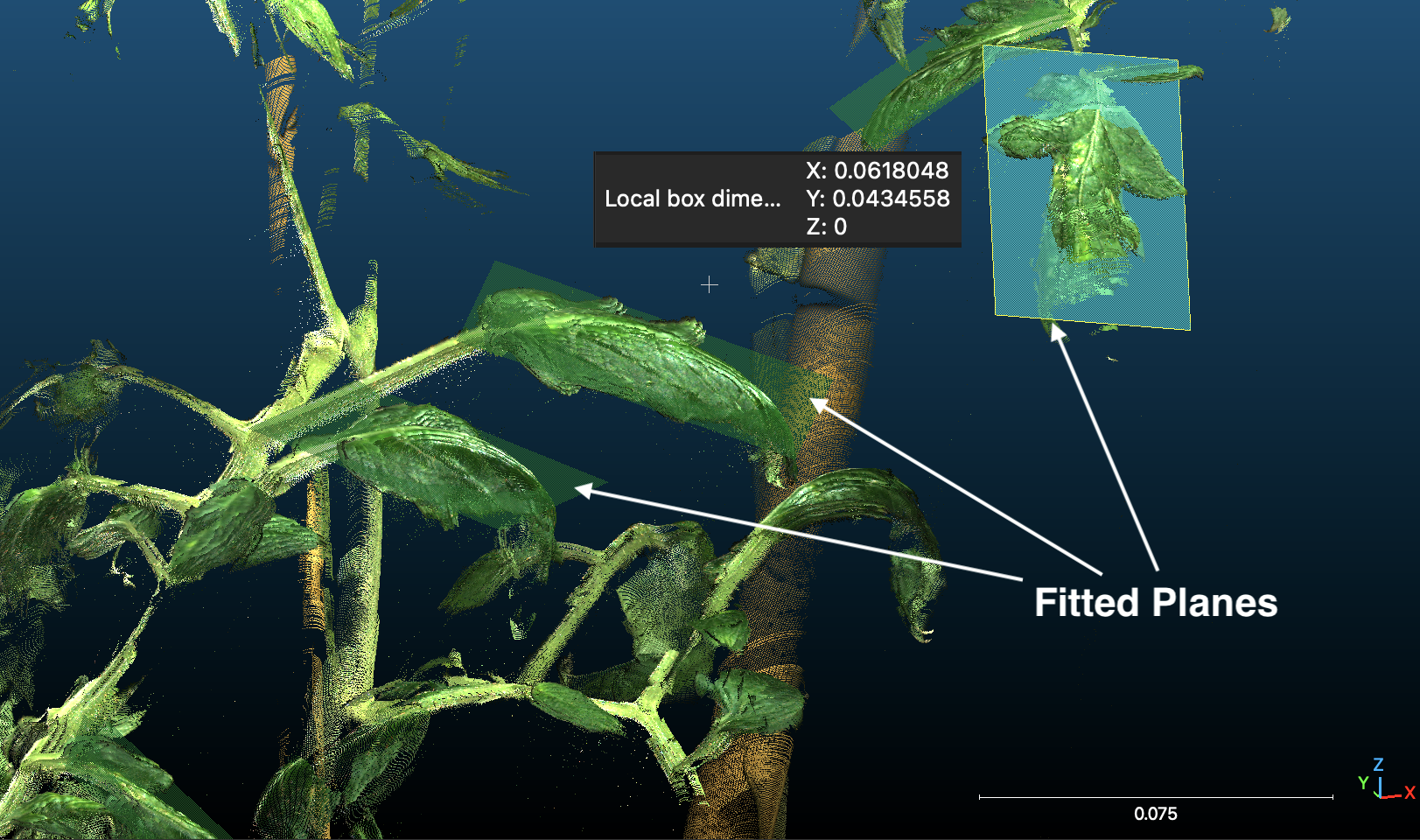}
            \caption{CloudCompare plane fitting}
            \label{fig:SoftwarePlaneFitting}
        \end{figure}

\begin{figure}[!htbp]
            \centering
            \includegraphics[width=75mm]{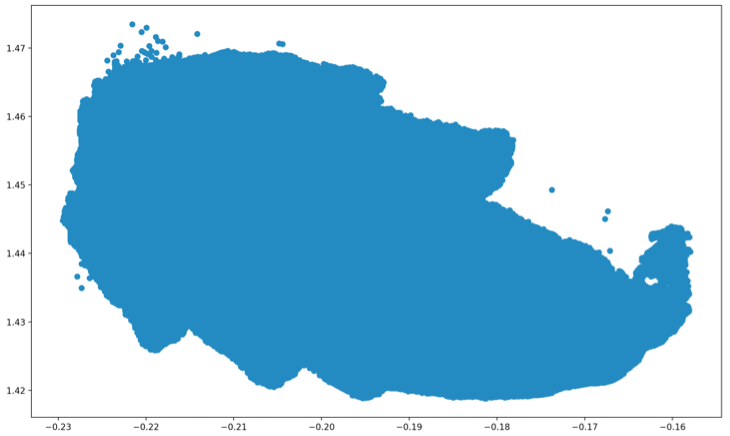}
            \caption{Algorithm plane fitting - 3D point projection on fitted plane.}
            \label{fig:AlgorithmPlaneFitting}
        \end{figure}

        As shown in Figure \ref{fig:SoftwarePlaneFitting}, Plane fitting was first tested using a CloudCompare function which can fit a plane on the point cloud of a leaf, giving the length and width of the plane, which enable tracking of the leaf size.   
        The length and width of the fitted plane show a high correlation to the measured leaf length and width. This method is then implemented in python using RANSAC plane fitting algorithms from two different libraries. Open3D's plane segmentation\footnote{\url{http://www.open3d.org/docs/latest/tutorial/Basic/pointcloud.html}} function and pyRANSAC-3D's plane function \footnote{\url{https://leomariga.github.io/pyRANSAC-3D/}} were used to output a plane equation of the fitted plane. Open3D is a widely used library for 3D data processing, and pyRANSAC-3D is a small library focusing on implementing the RANSAC method for various shapes. As shown in Figure \ref{fig:AlgorithmPlaneFitting}, the 3D leaf points are then projected onto the 2D plane to extract the length and width of the leaf on the plane. 
        
        This length and width correlate to the manually measured value and the software method, proving the algorithm is operational and can then be used to monitor the plant's growth. 
        
        The researched machine learning approach was not tested or implemented in this project due to its slow process of labelling and training. In contrast, the current approach can output the estimate in seconds. In addition lack of an extensive dataset for leaf size measurements also decreases the method's accuracy. This leads to a training dataset shortage, leading to inefficient and low-accuracy models. However, this creates room for future work with a larger dataset or synthetic data generated from existing data.
        
        For the following reasons, an overall assumption can be established for leaf size that the proposed method is robust against leaf curvature. Firstly, the length and width will remain the same across the manual measurement, software measurement and algorithm results due to consistent measuring methods. Secondly, for the proposed method, if curvature occurs, the leaf length will likely have an abnormal decrease, which can be used to draw special attention to care. Lastly, as this research aims to monitor the growth of the leaf, tracking the change in the length and width is sufficient for the aimed problem.

\section{Results}
This section presents the results of experiments done in testing and evaluating the proposed methods.

    \subsection{Data Collection}
    Table \ref{tab:cube_error} demonstrates the pre-existing error from the Zivid camera and the point cloud stitching algorithm error from the function used. The test was done with a 50mm cube being 300, 400, 500 and 600 mm away from the camera, which is the majority distance between the scanned leaf and the camera, and optimal working distance was recommended by Zivid. These point clouds were obtained using the same 50-degree view and combined twelve scans used in obtaining the Leaf-24 and Height-18 datasets.

        % Cube Error
        \begin{table}[]
        \caption{Camera and point cloud stitching error.}
        \begin{adjustbox}{width=\linewidth}
        \centering
        \begin{tabular}{|ccc|}
        \hline
        \multicolumn{1}{|c|}{Camera to cube(mm)} & \multicolumn{1}{c|}{Error Percentage (\%)} & RMSE (mm) \\ \hline
        \multicolumn{1}{|c|}{300}  & \multicolumn{1}{c|}{2.02} & 1.10 \\ \hline
        \multicolumn{1}{|c|}{400}  & \multicolumn{1}{c|}{1.27} & 0.83 \\ \hline
        \multicolumn{1}{|c|}{500}  & \multicolumn{1}{c|}{2.27} & 1.29 \\ \hline
        \multicolumn{1}{|c|}{600}  & \multicolumn{1}{c|}{3.8}  & 2.09 \\ \hline
        \end{tabular}
        \label{tab:cube_error}
        \end{adjustbox}
        \end{table}
    \subsection{Detection}
\begin{figure}[!htbp]
            \centering
            \begin{subfigure}[t]{42mm}
              \centering
              \includegraphics[width=42mm, height=42mm]{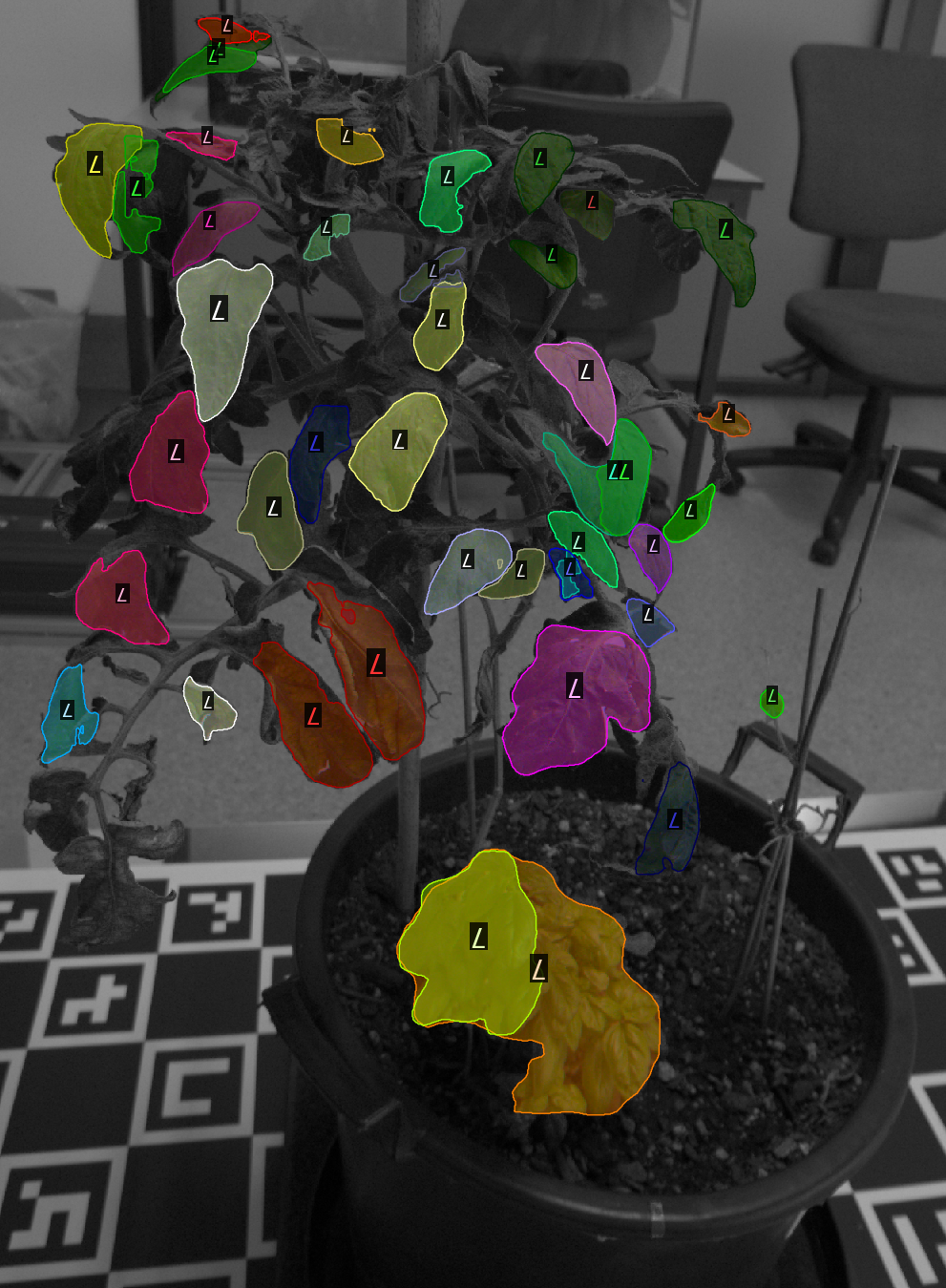}
              \caption{Poor}
              \label{fig:leafPredictions}
            \end{subfigure}
            \begin{subfigure}[t]{42mm}
              \centering
              \includegraphics[width=42mm, height=42mm]{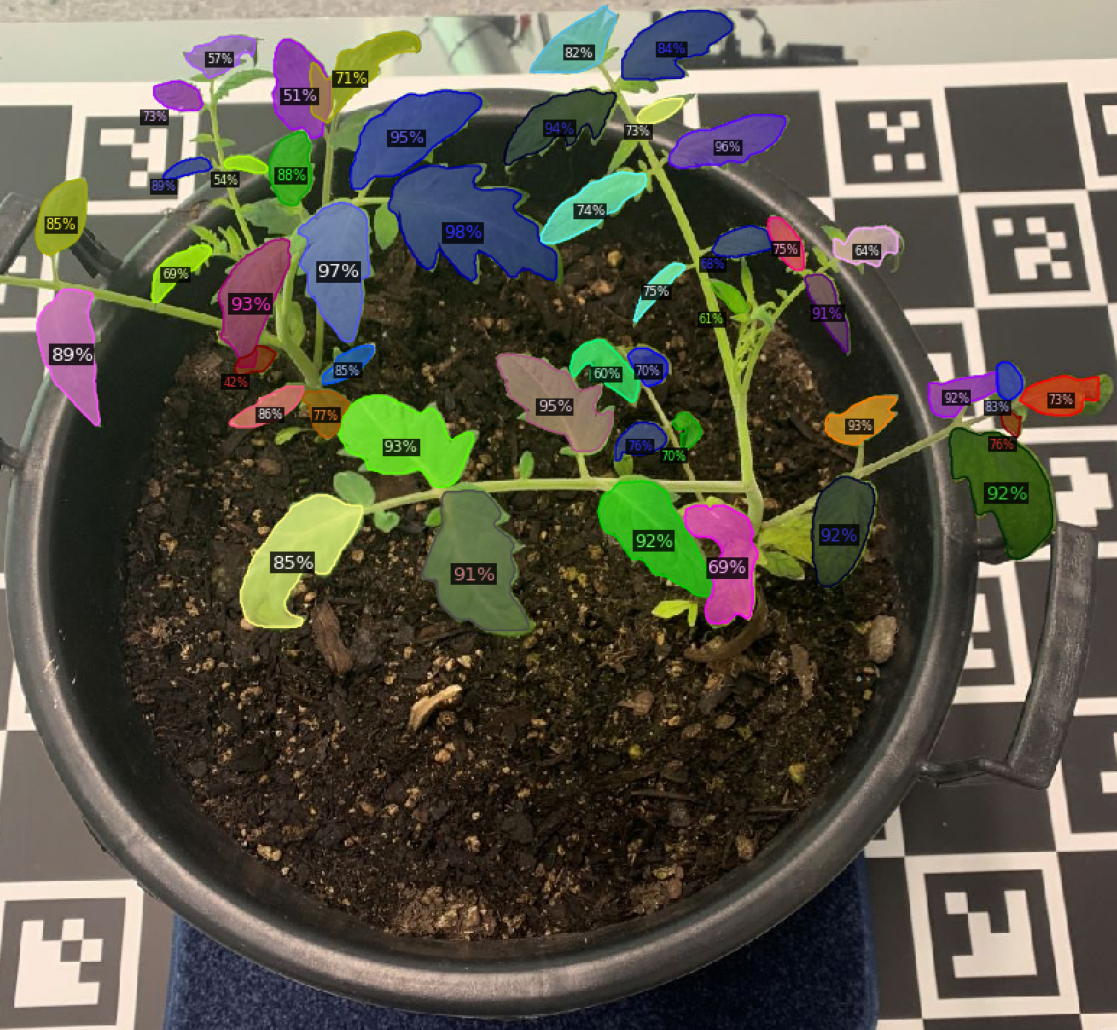}
              \caption{Reasonable}
              \label{fig:leafPredictions2}
            \end{subfigure}\\
            \caption[]{Result of leaf detection a)poor, b)reasonable.}
            \label{chap3:fig:kiwifruit_over_glare_typical}
        \end{figure}

    Detection Results shown in Table \ref{tab:detection_results} were evaluated using Detectron2's inbuilt evaluation method, which utilises the standard COCOEvaluator. Detection outputs are presented in Figure \ref{fig:leafPredictions} and \ref{fig:leafPredictions2}.
        % \begin{table}[h]
        % \caption{Leaf detection results.}
        % \begin{adjustbox}{width=\linewidth}
        % \centering
        % \begin{tabular}{|c||c|c|c|}
        
        % \hline
        % Training Method & AP & AP50 & AP75 \\ \hline
        % Mask R-CNN R50 fpn3x & 32.14 & 42.64 & 36.98 \\
        % \hline
        
        % % \begin{tabular}{|c|c|c|c|}

        % % % Training Method & AP & AP50 & AP75 \\ \hline
        % % % Mask R-CNN R50 fpn3x & 32.14 & 42.64 & 36.98 \\\hline
        % % \hline
        % % \multicolumn{1}{|c|}{Training Method} & \multicolumn{1}{c|}{AP} & \multicolumn{1}{c|}{AP50} & AP70 \\ \hline
        % % \multicolumn{1}{|c|}{Mask R-CNN R50 fpn3x}  & \multicolumn{1}{c|}{32.14} & \multicolumn{1}{c|}{42.64} & 36.98 \\ \hline

        % \multicolumn{4}{|c|}{
        %     \begin{tabular}[c]{@{}}
        %     *AP: Average Precision is calculated as the weighted mean of precision at different Intersection over Union threshold\end{tabular}} \\ \hline

        % % \end{tabular}

        % \end{tabular}
        % \label{tab:detection_results}
        % \end{adjustbox}
        % \end{table}

 % \begin{table}[h]
        % \caption{Leaf detection results.}
        % \begin{adjustbox}{width=\linewidth}
        % \centering
        % \begin{tabular}{|c||c|c|c|}

        \begin{table}[]
        \centering
        \caption{Leaf detection results.}
        \label{tab:detection_results}
        \begin{adjustbox}{width=\linewidth}
        \begin{tabular}{|lccc|}
        \hline
        \multicolumn{1}{|l|}{Training Method} &
          \multicolumn{1}{c|}{AP} &
          \multicolumn{1}{c|}{AP50} &
          AP75 \\ \hline
        \multicolumn{1}{|l|}{Mask R-CNN R50 fpn3x} &
          \multicolumn{1}{c|}{45.21} &
          \multicolumn{1}{c|}{53.64} &
          48.65 \\ \hline
        \multicolumn{4}{|l|}{\begin{tabular}[c]{@{}l@{}}*AP: Average Precision is calculated as the weighted mean \\ of precision at different Intersection over Union threshold\end{tabular}} \\ \hline
        \end{tabular}
        \end{adjustbox}
        \end{table}

    \subsection{Leaf Size Measurement}
        Leaf size results were produced by testing the methods on the Leaf-24 dataset to evaluate the accuracy of the proposed approach. The average length of the leaves in this dataset is around 60mm, and the average width is around 35mm Error percentage ($average error/average size$), RMSE and $R^2$ values of each proposed method are presented in Table \ref{tab:plane_fitting_result}. 
        The Table \ref{tab:plane_fitting_result}.a presents results calculated using manually measured values as ground truth, and Table \ref{tab:plane_fitting_result}.b uses software measurements as the ground truth. As the main goal of this project is to monitor the growth of the leaf, the relationship between the estimated value and absolute truth (measured value) is not as important, which leads to the results in Table \ref{tab:plane_fitting_result}.b being elevated and used. The software-estimated value may also be more accurate than the manually measured value, as manual measurement consists of human errors of incorrect eye estimate of the maximum distance or accidentally bending the leaf and changing its original dimension. 
        
\begin{figure}[!ht]
            \centering
            \includegraphics[width=80mm]{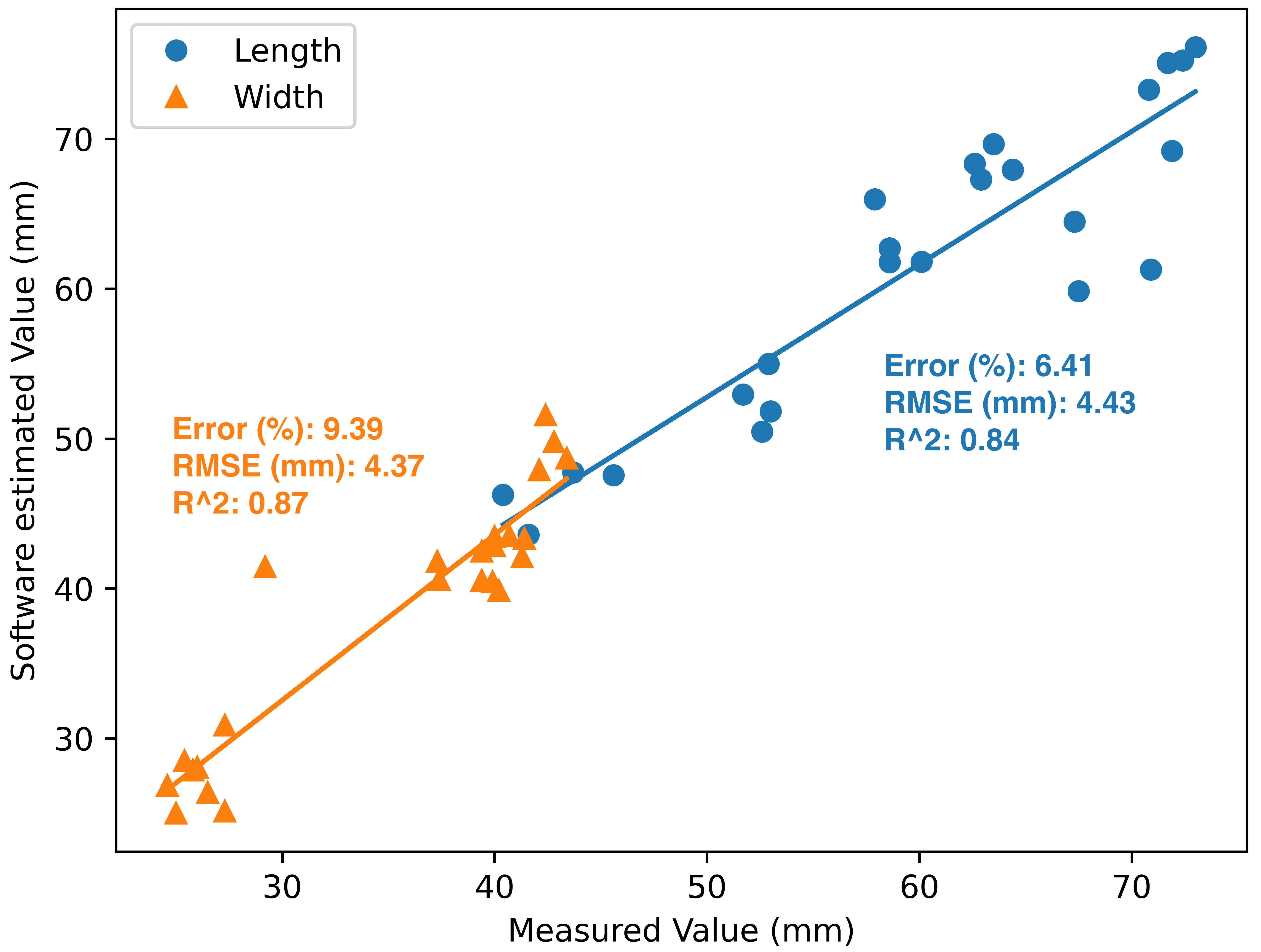}
            \caption{Leaf length accuracy comparison between measurement and software estimation for 6 leaves over 4 days.}
            \label{fig:groundTruths}
        \end{figure}

\begin{figure}[!ht]
            \centering
            \includegraphics[width=80mm]{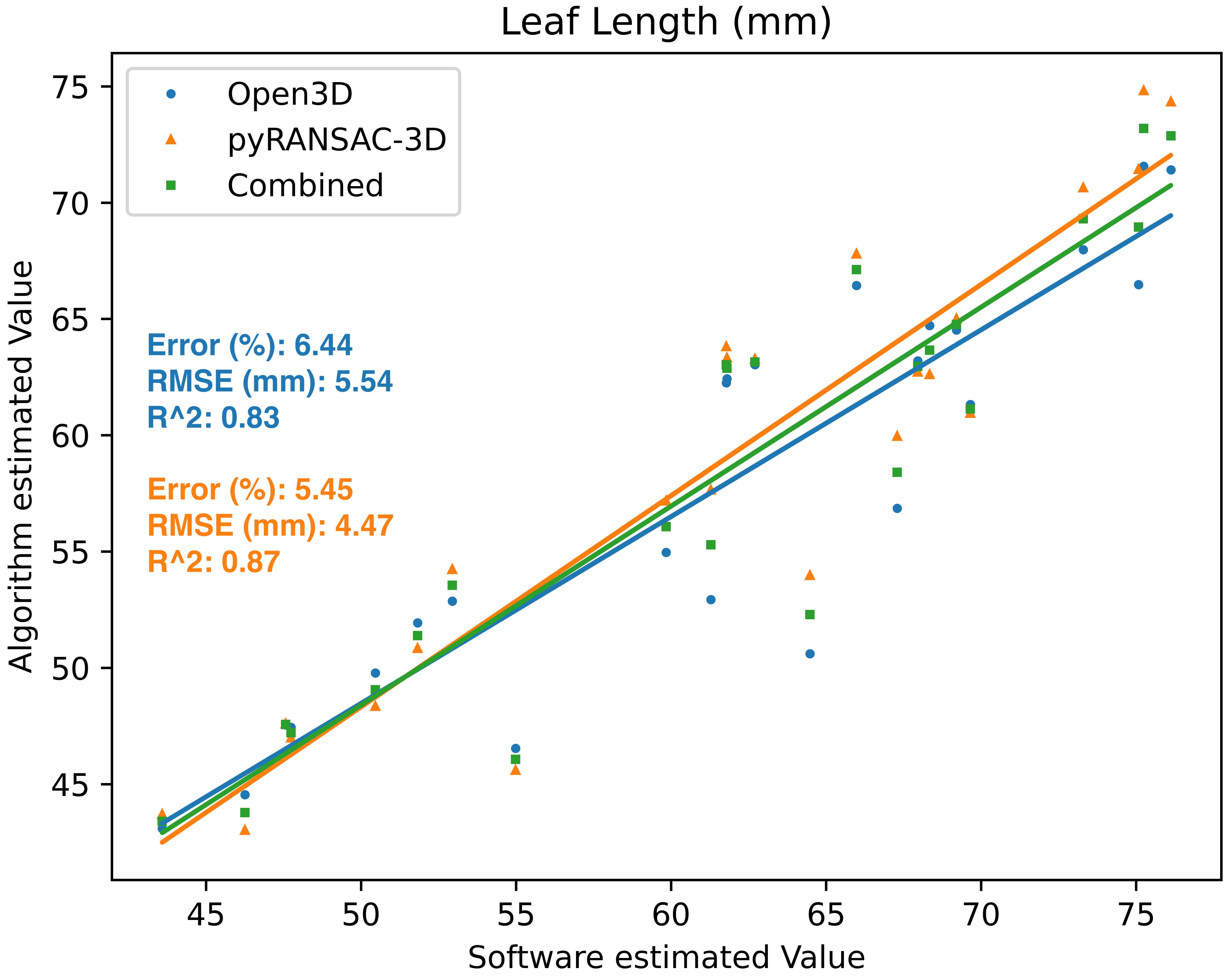}
            \caption{Leaf length accuracy comparison between software and algorithm estimation for 6 leaves over 4 days.}
            \label{fig:leafLength}
        \end{figure}

\begin{figure}[!ht]
            \centering
            \includegraphics[width=80mm]{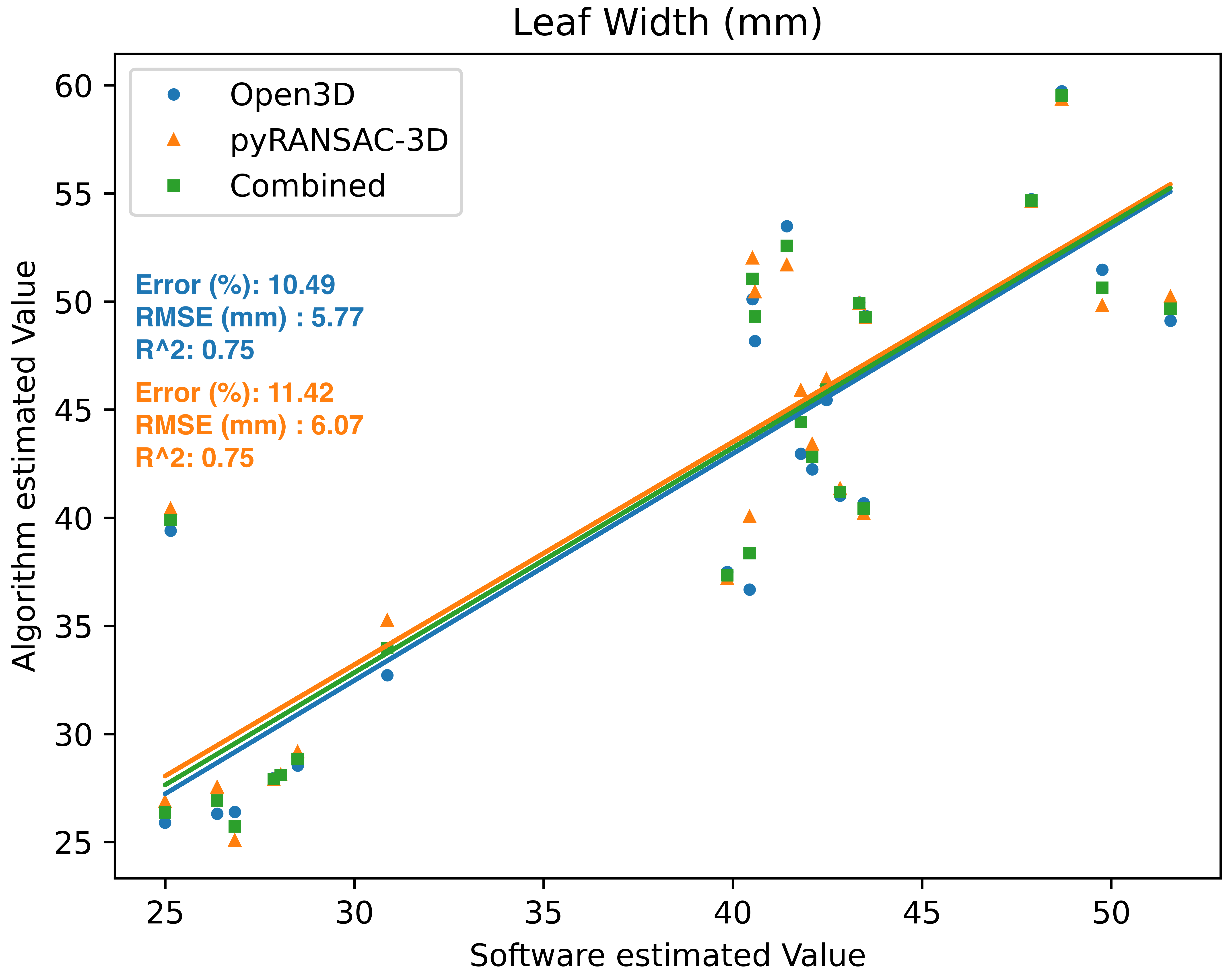}
            \caption{Leaf width accuracy comparison between software and algorithm estimation for 6 leaves over 4 days.}
            \label{fig:leafWidth}
        \end{figure}

\begin{figure}[!h]
            \centering
            \includegraphics[width=\linewidth]{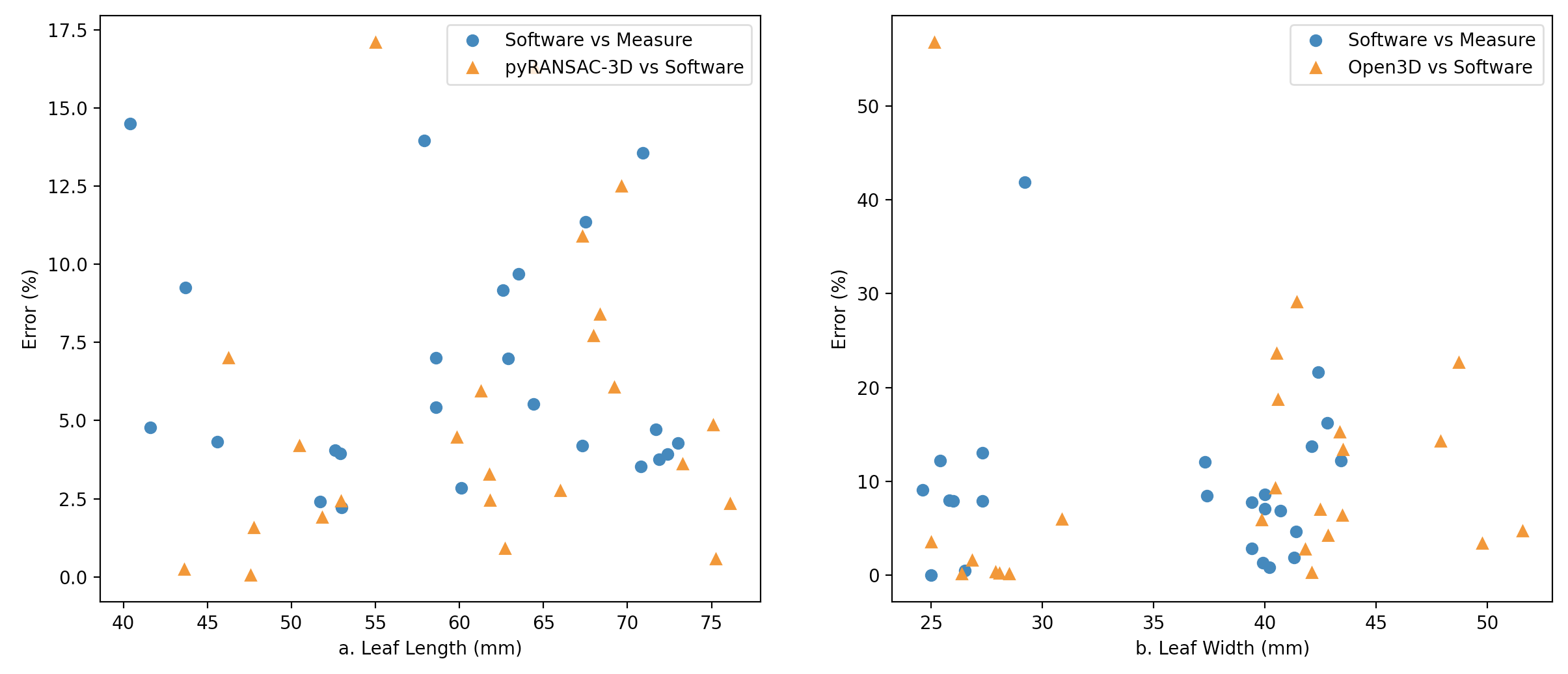}
            \caption{The relationship between leaf size (length(a), width(b)) and the percentage error.}
            \label{fig:leafError}
        \end{figure}

        To evaluate the relationship between the two ground truths, the accuracy of the software estimate and measured value in length and width are presented in Figure \ref{fig:groundTruths}, along with its percentage error, RMSE and $R^2$ value, which all suggest a linear relationship between the two. Followed by compression between software estimated value and algorithm estimated value for leaf length and width are presented in Figure \ref{fig:leafLength} and \ref{fig:leafWidth} respectively. Algorithm estimated values used the RANSAC plane fitting function from the previously mentioned libraries and combined the results, producing the highest result of RMSE=4.47mm and $R^2$= 0.87. Figure \ref{fig:leafError} presents the correlation between leaf size and percentage error to find any relation between the error and the size.
        
        \begin{table*}[h!]
        \caption{Plane fitting experiment results} 
        \centering
        \begin{adjustbox}{width=\textwidth}
        \begin{tabular}{ccccccccccc}
        \multicolumn{5}{c}{a. Manual measurement as ground truth} &
           &
          \multicolumn{5}{c}{b. Software measurement as ground truth} \\ \cline{1-5} \cline{7-11} 
        \multicolumn{1}{|c|}{} &
          \multicolumn{1}{c|}{} &
          \multicolumn{1}{c|}{\begin{tabular}[c]{@{}c@{}}Error \\ Percentage(\%)\end{tabular}} &
          \multicolumn{1}{c|}{RMSE(mm)} &
          \multicolumn{1}{c|}{$R^2$} &
          \multicolumn{1}{c|}{} &
          \multicolumn{1}{c|}{} &
          \multicolumn{1}{c|}{} &
          \multicolumn{1}{c|}{\begin{tabular}[c]{@{}c@{}}Error \\ Percentage(\%)\end{tabular}} &
          \multicolumn{1}{c|}{RMSE(mm)} &
          \multicolumn{1}{c|}{$R^2$} \\ \cline{1-5} \cline{7-11} 
        \multicolumn{1}{|c|}{\multirow{2}{*}{\begin{tabular}[c]{@{}c@{}}RANSAC - \\ Open3D\end{tabular}}} &
          \multicolumn{1}{c|}{Length} &
          \multicolumn{1}{c|}{8.24} &
          \multicolumn{1}{c|}{6.77} &
          \multicolumn{1}{c|}{0.60} &
          \multicolumn{1}{c|}{} &
          \multicolumn{1}{c|}{\multirow{2}{*}{\begin{tabular}[c]{@{}c@{}}RANSAC - \\ Open3D\end{tabular}}} &
          \multicolumn{1}{c|}{Length} &
          \multicolumn{1}{c|}{6.44} &
          \multicolumn{1}{c|}{5.54} &
          \multicolumn{1}{c|}{0.83} \\ \cline{2-5} \cline{8-11} 
        \multicolumn{1}{|c|}{} &
          \multicolumn{1}{c|}{Width} &
          \multicolumn{1}{c|}{18.39} &
          \multicolumn{1}{c|}{8.66} &
          \multicolumn{1}{c|}{0.62} &
          \multicolumn{1}{c|}{} &
          \multicolumn{1}{c|}{} &
          \multicolumn{1}{c|}{Width} &
          \multicolumn{1}{c|}{10.49} &
          \multicolumn{1}{c|}{5.77} &
          \multicolumn{1}{c|}{0.75} \\ \cline{1-5} \cline{7-11} 
        \multicolumn{1}{|c|}{\multirow{2}{*}{\begin{tabular}[c]{@{}c@{}}RANSAC - \\ pyRANSAC-3D\end{tabular}}} &
          \multicolumn{1}{c|}{Length} &
          \multicolumn{1}{c|}{7.27} &
          \multicolumn{1}{c|}{5.77} &
          \multicolumn{1}{c|}{0.69} &
          \multicolumn{1}{c|}{} &
          \multicolumn{1}{c|}{\multirow{2}{*}{\begin{tabular}[c]{@{}c@{}}RANSAC - \\ pyRANSAC-3D\end{tabular}}} &
          \multicolumn{1}{c|}{Length} &
          \multicolumn{1}{c|}{5.45} &
          \multicolumn{1}{c|}{4.47} &
          \multicolumn{1}{c|}{0.87} \\ \cline{2-5} \cline{8-11} 
        \multicolumn{1}{|c|}{} &
          \multicolumn{1}{c|}{Width} &
          \multicolumn{1}{c|}{19.31} &
          \multicolumn{1}{c|}{8.90} &
          \multicolumn{1}{c|}{0.64} &
          \multicolumn{1}{c|}{} &
          \multicolumn{1}{c|}{} &
          \multicolumn{1}{c|}{Width} &
          \multicolumn{1}{c|}{11.42} &
          \multicolumn{1}{c|}{6.07} &
          \multicolumn{1}{c|}{0.75} \\ \cline{1-5} \cline{7-11} 
          \multicolumn{1}{|c|}{\multirow{2}{*}{\begin{tabular}[c]{@{}c@{}}RANSAC - \\ Combined\end{tabular}}} &
          \multicolumn{1}{c|}{Length} &
          \multicolumn{1}{c|}{7.61} &
          \multicolumn{1}{c|}{6.19} &
          \multicolumn{1}{c|}{0.65} &
          \multicolumn{1}{c|}{} &
          \multicolumn{1}{c|}{\multirow{2}{*}{\begin{tabular}[c]{@{}c@{}}RANSAC - \\ Combined\end{tabular}}} &
          \multicolumn{1}{c|}{Length} &
          \multicolumn{1}{c|}{5.92} &
          \multicolumn{1}{c|}{4.91} &
          \multicolumn{1}{c|}{0.86} \\ \cline{2-5} \cline{8-11} 
        \multicolumn{1}{|c|}{} &
          \multicolumn{1}{c|}{Width} &
          \multicolumn{1}{c|}{18.81} &
          \multicolumn{1}{c|}{8.75} &
          \multicolumn{1}{c|}{0.63} &
          \multicolumn{1}{c|}{} &
          \multicolumn{1}{c|}{} &
          \multicolumn{1}{c|}{Width} &
          \multicolumn{1}{c|}{10.95} &
          \multicolumn{1}{c|}{5.88} &
          \multicolumn{1}{c|}{0.75} \\ \cline{1-5} \cline{7-11} 
        \multicolumn{1}{|c|}{\multirow{2}{*}{\begin{tabular}[c]{@{}c@{}}CloudCompare\end{tabular}}} &
          \multicolumn{1}{c|}{Length} &
          \multicolumn{1}{c|}{6.41} &
          \multicolumn{1}{c|}{4.43} &
          \multicolumn{1}{c|}{0.84} &
          \multicolumn{1}{c|}{} &
          \multicolumn{1}{c|}{\multirow{2}{*}{\begin{tabular}[c]{@{}c@{}}Overall \\ Pipeline\end{tabular}}} &
          \multicolumn{1}{c|}{Length} &
          \multicolumn{1}{c|}{13.9} &
          \multicolumn{1}{c|}{9.77} &
          \multicolumn{1}{c|}{0.899} \\ \cline{2-5} \cline{8-11} 
        \multicolumn{1}{|c|}{} &
          \multicolumn{1}{c|}{Width} &
          \multicolumn{1}{c|}{9.39} &
          \multicolumn{1}{c|}{4.37} &
          \multicolumn{1}{c|}{0.87} &
          \multicolumn{1}{c|}{} &
          \multicolumn{1}{c|}{} &
          \multicolumn{1}{c|}{Width} &
          \multicolumn{1}{c|}{16.4} &
          \multicolumn{1}{c|}{8.13} &
          \multicolumn{1}{c|}{0.68} \\ \cline{1-5} \cline{7-11} 
          
        %   \multicolumn{1}{|c|}{\multirow{2}{*}{\begin{tabular}[c]{@{}c@{}}CloudCompare\end{tabular}}}  &
        %   \multicolumn{1}{c|}{Length} &
        %   \multicolumn{1}{c|}{6.41} &
        %   \multicolumn{1}{c|}{4.43} &
        %   \multicolumn{1}{c|}{0.84} \\ \cline{2-5} \cline{8-11} 
        % %   &
        % %   &
        % %   &
        % %   &
        % %   \\ \cline{1-5}
        
        %   \multicolumn{1}{c|}{Width} &
        %   \multicolumn{1}{c|}{9.39} &
        %   \multicolumn{1}{c|}{4.37} &
        %   \multicolumn{1}{c|}{0.87} &
        %   &
        %   &
        %   &
        %   &
        %   \\ \cline{1-5}

        \end{tabular}
        \label{tab:plane_fitting_result}
        \end{adjustbox}
        \end{table*}

        \subsection{Integration of detection and leaf size.}
        The overall performance of the integrated measurement system is presented in Table \ref{tab:plane_fitting_result}. 

\section{Discussion}
The overall proposed system achieved an RMSE value of 8.13mm and an $R^2$ of 0.899 for leaf size detection. This is due to the combination of errors from detection and size measurement. Errors also exist with the alignment between RGB and point cloud when mapping the 2D leaf mask to its corresponding 3D point cloud. This section focuses on further explaining the source of these errors that led to the final result.

    \subsection{Data Collection}
    Camera and point cloud stitching errors exist, as shown in Table \ref{tab:cube_error}, from a controlled experiment with minimum human measurement error. As both percentage error and RMSE display, accuracy rises as the distance is shortened between the camera and the object. However, a higher error occurred at 300mm, due to the insufficient data collected as it is too close to the camera. Using software estimated value as ground truth bypasses this error as both software and algorithm work on the same point cloud. This can be further proven by the reduced error of around 2\% in length between the two ground truths. These errors are likely caused by the noise created when combining all 12 scans from different angles using the pre-existing point cloud stitching algorithm. In addition, minor camera errors and object stability may also contribute to the errors in producing the point cloud. However, with all these in mind, a maximum of 3.8\% error and 2.09mm RMSE value can be seen as acceptable and as the foundation error existing in the point cloud, which can be taken away from the final result of the proposed system in order to produce a more accurate result. Further experiments can be done using different size objects to assist in confirming this error.

    \subsection{Detection}
        The overall experimental results produced by the Mask R-CNN model were reasonable, with an AP value of 45.21, this number increases to 53.64 considering AP50. It can be observed that the performance varies in Figure \ref{fig:leafPredictions} and \ref{fig:leafPredictions2}. The leaf density would be changed during the plant growth stages. A dense plant like Figure \ref{fig:leafPredictions2} would cause leaf occlusion and it resulted in poor performance. In general, plants are complex organisms that dynamically change in size and structure at different stages of their life cycles \cite{minervini2015image}. The results were partly expected when conducting this research, as no publicly available dataset for tomato plants exists. However, different data augmentation techniques were implemented to increase the available dataset size. It seems that more annotated data is required.
        
        %Likely, the annotated dataset could only contain tomato leaves at a specific growth stage as plants, in general, are complex organisms that dynamically change in size and structure at different stages of their life cycles \cite{minervini2015image}.
        
        %It is worth pointing out that in some occurrences, when the tomato plants are small and in their early growth stages, the trained model could detect a majority of leaves, such as the one shown in Figure \ref{fig:leafPredictions2}. Although the reason for this was unclear, an educated guess is that tomato leaves might be easier to detect in some growth stages, and they might require less annotated data to train the model efficiently. 

    \subsection{Leaf Size Measurement}
     Higher accuracy of RMSE=4.47mm and $R^2$= 0.87 were achieved when having software estimated value as ground truth, as shown in Table \ref{tab:plane_fitting_result}. This is likely due to the reduced human error in measuring the leaf and the high correlation between local length and width difference due to manual measurement issues. Camera and point cloud stitching errors also reduce the accuracy, which will be further discussed later. Software estimated value can be used as ground truth in this project as Figure \ref{fig:groundTruths} shows its linear relationship with the measured value, and a high $R^2$ value of 0.87 and low RMSE of 4.37mm. Three different sources likely cause the difference between these results. Firstly human error on the measured value side; when manually measuring with a calliper, the maximum length and width are slightly inaccurate due to human eye estimation. Secondly, software plane fitting accuracy may vary when cropping out the individual leaves when fitting a plane. Lastly, the point cloud of the tomato leaves likely contains stitching errors (as presented in Section 4.1) when combining twelve individual files, resulting in the misalignment between the digital twin and the actual plant. From Figure \ref{fig:leafLength} and \ref{fig:leafWidth}, linear relationships can be seen between software and algorithm estimated value. 
        
    pyRANSAC-3D library's plane fitting function produces a better result in estimating the length with RMSE = 4.47mm and $R^2$ = 0.87. On the other hand, the Open3D library's plane fitting function produces a better width result with RMSE = 6.07mm and $R^2$ = 0.75. As can be observed in Table \ref{tab:plane_fitting_result}, width has a higher error than the length in general. A higher error difference occurs when manual measurement as ground truth is due to the eye estimate error in measuring width, which is more inaccurate than length due to the leaf shape. As presented in Figure \ref{fig:leafError}, no correlation between the error and the leaf size shows its consistency. As the leaf size varies dramatically throughout the plant's growing process, this stability is crucial to producing accurate data for growth monitoring.

    As non-destructive automatic leaf size measurement was not extensively researched previously, this result could not be directly compared with other approaches. Looking at a destructive and highly controlled environment approach, Nyakwende et al. \cite{nyakwende1997non} achieved a $R^2$ value of 0.992. While not completely comparable, other 3D data techniques used to measure fruit-let diameter \cite{wang2014size} (RMSE = 3.4mm, $R^2$ = 0.89) and leaf area \cite{zhang20183d} (RMSE = 0.12 $m^2$, $R^2$ = 0.98) also testify to the common error of 3D approaches. Considering the robustness, non-destructive approach and the detailed shape of the singular leaf, the 0.87 $R^2$ value and 4.47 mm RMSE achieved in this paper can be considered a reasonably accurate result with room for improvement.
    
    Additionally, the difference in predicted value and the actual value is not important, whether it is consistent or not. As we are monitoring the tomato plants over their growth period, which have rapidly growing leaves, the minor error would not affect the monitoring due to the expected significant change between measurements. The leaf movements also caused this difference during manual measurement and compared to the static leaf during software measurement; there will be unmeaningful errors that do not affect the result.
        
    Given the results analyzed, the proposed approach will use the pyRANSAC-3D plane function to estimate the length of the leaf and the Open3D plane function to estimate the width of the leaf. 

\section{Conclusions and Future Work}
    % Example text you can adjust to the specific task for this paper
    
    This paper presents the initial design and integration of an automated non-destructive tomato plant growth system. Unlike previous destructive leaf measuring methods, specific attempts have been made to measure and extract the size information without destroying the plant. The monitoring system was evaluated on 24 scans of tomato plants in an indoor lab under lab conditions. The leaf size measurement results demonstrate that the algorithm can measure tomato leaf size with an RMSE of 4.47mm and an $R^2$ of 0.87, while the overall results achieve an RMSE of 9.77mm and an $R^2$ of 0.899.
    
    At the current state, the detection pipeline is the main bottleneck of the proposed monitoring system. Future work is still required to reduce the under-counting through improvements to the detection system and further developments to the plane fitting approach. In some scenarios, the model can still detect a majority of tomato leaves in unseen images. This is likely because the leaves in such images have simple texture and geometry due to being at early growth stages. However, further research in this area with a larger dataset size is required before coming up with a conclusion. 
    
    Going forward, experiments can be done to improve the algorithm leaf size measurement accuracy by denoising the point cloud and increasing the number of individual scans per point cloud, including scans from different sides of the plant. Machine Learning algorithms for leaf size estimation can also be implemented as the dataset increases. Additionally, the plant height dataset can be further tested to implement plant height measurement, increasing the robustness and accuracy of the growth monitoring with the additional feature. Plant height algorithms can also be tested to evaluate their accuracy in finding the distance between vine and soil or the top plane to the bottom plane. The size of the open-source dataset will also increase as more scans and measurements are done.

\section*{Acknowledgements}
    This research was funded by the New Zealand Ministry for Business, Innovation and Employment (MBIE) on contract UOAX2116, an Artificial Intelligence-based Smart Farming System. Mahla Najati is the corresponding author. 
    
\bibliography{publications}
\bibliographystyle{named}

\end{document}